\documentclass{article}
\usepackage{spconf,amsmath,graphicx,adjustbox,bm,booktabs,hyperref}
\usepackage{cite}
\usepackage{mymacros}

\usepackage{bm, color}
\usepackage{amsmath}
\usepackage{amsfonts}
\title{Relational Future Captioning Model \\ for Explaining Likely Collisions in Daily Tasks
}
\name{Motonari Kambara and Komei Sugiura
\thanks{This material is based upon work supported by JSPS KAKENHI Grant Number 20H04269, JST Moonshot, and NEDO.}
}
\address{Keio University, Japan
}
\begin{document}
\setlength{\abovedisplayskip}{2pt} 
\setlength{\belowdisplayskip}{2pt} 
%
\maketitle
\begin{abstract}
Domestic service robots that support daily tasks are a promising solution for elderly or disabled people. It is crucial for domestic service robots to explain the collision risk before they perform actions. In this paper, our aim is to generate a caption about a future event. 
We propose the Relational Future Captioning Model (RFCM), a crossmodal language generation model for the future captioning task. The RFCM has the Relational Self-Attention Encoder to extract the relationships between events more effectively than the conventional self-attention in transformers. We conducted comparison experiments, and the results show the RFCM outperforms a baseline method on two datasets.
\end{abstract}
\begin{keywords}
Future Captioning, Domestic Service Robots, Relational Self-Attention
\end{keywords}
\vspace{-4mm}
\section{Introduction}
\vspace{-3mm}
\label{intro}


Domestic service robots (DSRs) that naturally communicate with users to support household tasks are a promising solution for elderly or disabled people. DSRs are expected to perform most tasks autonomously, and so they could damage objects and themselves. It therefore would be useful if they could explain the potential risks associated with their actions through natural language. However, a DSR's ability to generate natural language explanations is still insufficient.

\begin{figure}[t]
    \centering
    \includegraphics[height=45mm]{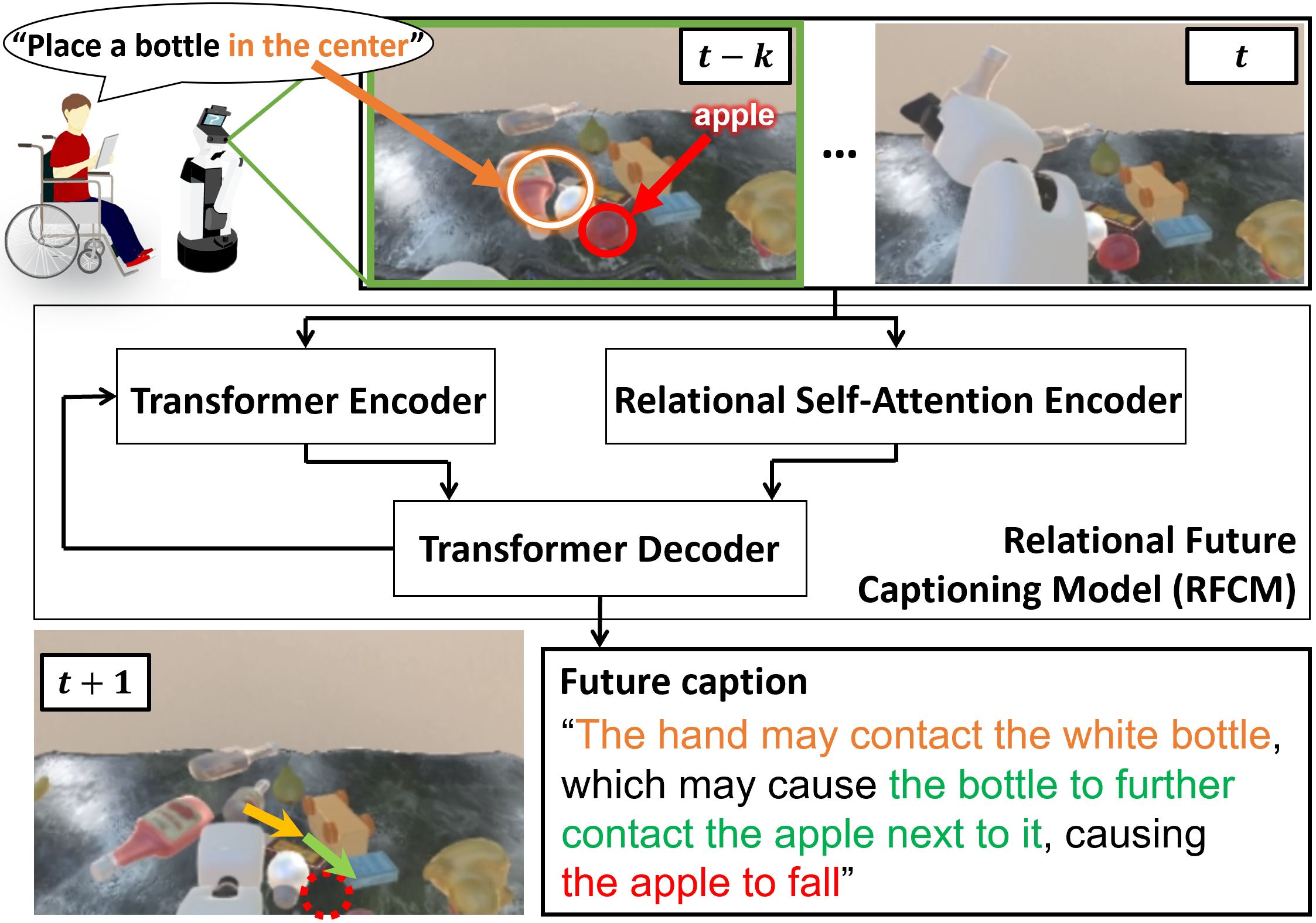}
    \vspace{-5mm}
    \caption{\small Overview of RFCM: RFCM generates captions of future events from past events.}
    \vspace{-7mm}
    \label{fig:eye_catch}
\end{figure}

Given this background, we focus on future captioning for daily tasks~\cite{hosseinzadeh2021video}. Fig.~\ref{fig:eye_catch} shows a typical use case where a DSR puts a plastic bottle on a table. In this situation, it would be desirable to tell the user that ``the hand may contact the white bottle, which may cause the bottle to further contact the apple next to it, causing the apple to fall.'' This task is difficult in that models need to predict future events and generate captions. In fact, there is a big gap in the quality of the reference and the generated sentences by typical video captioning models, as shown in Sec.~\ref{exp}.

Although there are many existing video captioning models, those models  cannot generate appropriate future captions. This is because many existing methods insufficiently model the temporal relationship between the visual features and sentences. The transformer self-attention~\cite{vaswani2017attention} used in those methods typically models the relationship between the visual features of the event and the sentence at time $t$. In addition, many video captioning methods are inappropriate for future captioning because those models use the captions after the next timestep to generate a caption for the next timestep.

In this paper, we propose the Relational Future Captioning Model (RFCM). The RFCM can generate captions that take into account the relationship between past events. This is because it has a source--target attention structure that generates appropriate captions for future events from the relationships between events. In this structure, the features derived from past clips are used as a source, and the features derived from both past clips and captions are used as the target. 

Fig.~\ref{fig:eye_catch} shows an overview of the RFCM, which consists of three modules: a Relational Self-Attention (RSA~\cite{kim2021relational}) Encoder, transformer encoder, and transformer decoder. The difference with respect to existing methods is that our method includes the RSA Encoder to more effectively extract the relationships between events than the conventional self-attention in transformers. Our code is available at this URL\footnote{\url{https://github.com/keio-smilab22/RelationalFutureCaptioningModel}}.


The main contributions of this paper are as follows:
\vspace{-2mm}
\begin{itemize}
  \setlength{\parskip}{-0.2mm}
  \setlength{\itemsep}{-0.5mm}
 \item We propose the RFCM, a crossmodal language generation model for the future captioning task. 
 \item The RSA Encoder is introduced to extract the relationships between events more effectively than the conventional self-attention in transformers. The output of the RSA Encoder is used to compute a source--target attention to obtain fine-grained caption representations. 
\end{itemize}
\vspace{-6mm}
\section{Related Work}
\vspace{-4mm}
\label{related}
There have been many studies in the field of caption generation~\cite{wang2018video, krishna2017dense, xu2015show, lei2020mart, hosseinzadeh2021video, kambara2021case, magassouba2021predicting}. This field includes image captioning and video captioning. \cite{hossain2019comprehensive} is a survey paper in the image captioning field. 
\cite{aafaq2019video} is a survey paper in the video captioning field. 

The field of video captioning can be divided into several subtasks such as a video captioning task, dense video captioning task, and future captioning task. The video captioning task involves generating a description about an input event. Many video captioning models have been proposed~\cite{zhao2021multi, wang2018video, lei2020mart, sun2019videobert, luo2020univl}.
\cite{sun2019videobert, luo2020univl} are typical examples of the numerous pre-training models. 
The dense video captioning task involves the generation of detailed captions.
\cite{deng2021sketch, mun2019streamlined, krishna2017dense} are representative dense video captioning models.
The future captioning is the task dealt with in this paper.
\cite{hosseinzadeh2021video, moriy2021, mahmud2021103230} are representative future captioning models. 
\cite{moriy2021} generates explanatory sentences of near future events using in-vehicle camera images and vehicle motion information observed from past to present.
\cite{hosseinzadeh2021video} and \cite{moriy2021} used the CNN and LSTM encoder-decoder, while our method used the transformer encoder-decoder.
\cite{lei2020more} and \cite{wu2021greedy} proposed event and video prediciton models, respectively. 
\vspace{-6mm}
\section{Problem Statement}
\vspace{-4mm}
\label{sec:problem}



Our target is to generate a sentence describing the next event in the video, which we call the future captioning task~\cite{hosseinzadeh2021video}. Fig.~\ref{fig:eye_catch} shows a typical scene. In this task, the desired behavior is to generate a description by inferring the situation at time $t+1$ given the clips up to time $t$.

The task is characterized by the following:
\vspace{-2mm}
\begin{itemize}
  \setlength{\parskip}{-0.2mm}
  \setlength{\itemsep}{-0.5mm}
    \item \textbf{Input:} video clips up to time $t$. We define the term ``clip'' as a sequence of frames representing an event.
    \item \textbf{Output:} a sentence describing an event that is expected to occur at time $t+1$.
\end{itemize}
\vspace{-2mm}
We assume that pretraining on other tasks is not allowed because knowledge transfer is outside the scope of this study.

%
%
%
\vspace{-6mm}
\section{Proposed Method}
\label{method}
\vspace{-4mm}
\begin{figure}
    \centering
    \includegraphics[width=\linewidth]{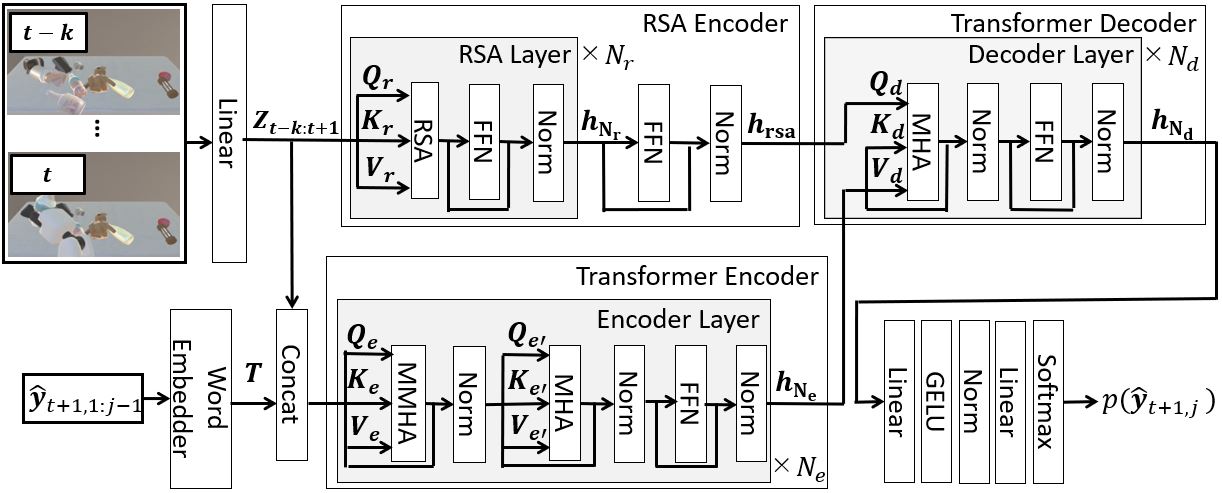}
    \vspace{-5mm}
    \caption{\small The framework of RFCM. In the figure, Norm denotes layer normalization.}
    \vspace{-7mm}
    \label{fig:encoder-decoder}
\end{figure}

%

Fig.~\ref{fig:encoder-decoder} shows the framework of the RFCM. 
The input to the RFCM $\bm{x} \in \mathbb{R}^{(k+1) \times d_{\mathrm{in}}}$ is defined as $\bm{x} = \{\bm{x}_{t-k}, ..., \bm{x}_t\},$ where $\bm{x}_t \in \mathbb{R}^{d_{\mathrm{in}}}$ and $k \in \mathbb{N}$ denote the clip at time $t$ and the number of input events that occurred before $t$, respectively.

From $\bm{x}$, $\bm{Z}_{t-k:t+1} \in \mathbb{R}^{(k+2) \times d_{\mathrm{in}}}$ is obtained as follows:
\begin{align}
    \bm{z}_{\tau} &= \left \{
        \begin{array}{l}
            f_z(\bm{x}_{\tau}) \quad (\tau > t)  \nonumber \\
            f_z(\bm{x}_{\tau}, \bm{x}_t) \quad (\tau \leq t), \nonumber
        \end{array}
    \right. \\
    \bm{Z}_{t-k:t+1} &= \{\bm{z}_{t-k}; ...; \bm{z}_{t+1}\}, \nonumber 
\vspace{-0.5mm}
\end{align}
where $\tau = t-k, ..., t+1$, $f_z(\cdot)$ denotes a linear transformation. $\bm{Z}_{t-k:t+1}$ contains information on an event sequence.
\vspace{-8mm}
\subsection{RSA Encoder}
\vspace{-2mm}
The RSA Encoder uses RSA~\cite{kim2021relational}, which
uses correlations between query and key, and self-correlations among values as the relational context. Although simple source-target attention
does not use self-correlations among values,  
it is important to consider their correlations because video is a time-series data.

The RSA Encoder consists of $N_r$ RSA layers. In the first layer, positional encoding using trigonometric functions is performed for $\bm{Z}_{t-k:t+1}$ using the procedure described in~\cite{vaswani2017attention}.

Then, query $\bm{Q}_r\in \mathbb{R}^{d_{\mathrm{rsa}}}$, key $\bm{K}_r \in \mathbb{R}^{(k+2) \times d_{\mathrm{rsa}}}$, and value $\bm{V}_r \in \mathbb{R}^{(k+2) \times d_{\mathrm{rsa}}}$ are obtained as follows:
\vspace{-1mm}
\begin{align}
\bm{Q}_r = \bm{z}_t, \bm{K}_r = \bm{V}_r = \bm{Z}_{t-k:t+1}, \nonumber
\vspace{-2mm}
\end{align}
where $d_{\mathrm{rsa}}$ denotes the size of each layer. Next, the basic kernel $\bm{\varphi}_p \in \mathbb{R}^{k+2}$ and relational kernel $\bm{\varphi} _h \in \mathbb{R}^{k+2}$ are obtained as follows:
\vspace{-1mm}
\begin{align}
\bm{\varphi}_p &= \bm{W}_p \bm{Q}_r, \nonumber \\
\bm{Q}_{r'} &= \{ \bm{Q}_r; ...;\bm{Q}_r\} \in \mathbb{R}^{(k+2) \times d_{\mathrm{rsa}}}, \nonumber \\ 
\bm{\varphi}_h &= \bm{W}_h f_{\mathrm{flatten}}(\bm{Q}_{r'} \odot \bm{K}_r),  \nonumber
\end{align}
where $f_{\mathrm{flatten}}(\cdot)$ denotes flattening. Next, relational context $\bm{\varPhi}_g \in \mathbb{R} ^ {(k + 2) \times d_{\mathrm{rsa}}}$ obtained as follows:
\begin{align}
\bm{\varPhi}_g &= \bm{V}_r + \bm{W}_g \bm{V}_r^\top \bm{V}_r. \nonumber
\vspace{-2mm}
\end{align}
Finally, RSA $\bm{\varphi} \in \mathbb{R} ^ {d_{\mathrm{rsa}}}$ is obtained as follows:
\begin{align}
\bm{\varphi} &= (\bm{\varphi}_p + \bm{\varphi}_h)^\top \bm{\varPhi}_g. \nonumber
\vspace{-2mm}
\end{align}
RSA $\bm{\varphi}$ contains information about the relationships between input events. To use $\bm{\varphi}$ as a latent feature at time $t$, we replace $\bm{z}_t$ with $\bm{\varphi}$. From this, we obtain $\bm{h}_r = \{ \bm{z}_{t-k}; ...;\bm{z}_{t-1}; \bm{\varphi}; \bm{z}_{t+1}\} \in \mathbb{R}^{(k+2) \times d_{\mathrm{rsa}}}$. The output of layer $\bm{h}_{\mathrm{n_r}}$ ($n_r = 1, ..., N_r$) is obtained by applying a feedforward network (FFN) and layer normalization (LN) layers to $\bm{h}_r$. 

The output of the encoder $\bm{h}_{\mathrm{rsa}} \in \mathbb{R}^{(k+2) \times d_{\mathrm{rsa}}}$ is given by $\bm{h}_{rsa} = f_{\mathrm{LN}}(f_{\mathrm{FFN}}(\bm{h}_{\mathrm{N_r}})),$ where $f_{\mathrm{FFN}}(\cdot)$ and $f_{\mathrm{LN}}(\cdot)$ denote the FFN and LN layers, respectively.
%
\vspace{-4mm}
\subsection{Transformer encoder/decoder}
\label{subsubsec:trmenc}
\vspace{-2mm}
The transformer encoder consists of $N_e$ encoder layers. Each layer consists of a Masked Multi-Head Attention (MMHA), Multi-Head Attention (MHA), and FFN layers.

The input to the encoder is $\bm{h}_c = \{\bm{Z}_{t-k:t+1}; \bm{T}\} \in \mathbb{R}^{(k+I+2) \times d_{\mathrm{in}}}$, where $I$ denotes the maximum length of sentences. During training, $\bm{T} \in \mathbb{R}^{I \times d_{\mathrm{in}}}$ denotes the text features obtained by embedding a reference sentence at time $t+1$, $\bm{y}_{t+1}$ with the BERT embedder~\cite{devlin2018bert}. 
During inference, $\bm{T}$ denotes the text features for the generated words $\hat{\bm{y}}_{t+1, 1:j-1}$ when the $j$-th word is generated. 


The input to the $n_e+1$-th layer is $\bm{h}_{n_e}$ ($n_e = 0,...,N_e-1$). We set $\bm{h}_0 = \bm{h}_c$. In the MMHA layer, query $\bm{Q}_e \in \mathbb{R}^{(k+I+2)\times d_e}$, key $\bm{K}_e \in \mathbb{R}^{(k+I+2)\times d_e}$, and value $\bm{V}_e \in \mathbb{R}^{(k+I+2)\times d_e}$ ($d_e = d_{\mathrm{enc}}/N_h$) are obtained as $\bm{Q}_e = \bm{W}^{(e)}_q \bm{h}_{n_e}$, $\bm{K}_e = \bm{W}^{(e)}_k \bm{h}_{n_e}$, and $\bm{V}_e = \bm{W}^{(e)}_v \bm{h}_{n_e}$($e = 1,...,N_h$), where $N_h$ and $d_{\mathrm{enc}}$ denote the number of attention heads and the size of each layer, respectively. During training, in the MMHA layer, we mask the $m$-th and subsequent word tokens to prevent the encoder from using the information of the words in the $m$-th word prediction (teacher forcing). To obtain attention $\bm{A}_{\mathrm{MMHA}} \in \mathbb{R}^{(k+I+2)\times d_\mathrm{enc}}$ from $\bm{Q}_e, \bm{K}_e$, and $\bm{V}_e$, we used the computation shown in~\cite{vaswani2017attention}.
Then, in the MHA layer, attention $\bm{A}_{\mathrm{MHA}}$ is obtained using the same computation used in the MMHA layer, where $\bm{h}_{\mathrm{enc}}$, $\bm{Q}_e$, $\bm{K}_e$, and $\bm{V}_e$ are replaced by $\bm{A}_{\mathrm{MMHA}}$, $\bm{Q}_{e'}$, $\bm{K}_{e'}$, and $\bm{V}_{e'}$, respectively. The output of the $n_e+1$-th encoder layer, $\bm{h}_{n_e+1} \in \mathbb{R}^{(k+I+2) \times d_{\mathrm{enc}}}$, is given by $\bm{h}_{n_e+1} = f_{\mathrm{LN}}(f_{\mathrm{FFN}}(\bm{A}_{\mathrm{MHA}})).$
The output of the encoder is $\bm{h}_{N_e}$.
%

The transformer decoder takes as inputs $\bm{h}_{\mathrm{rsa}}$ and $\bm{h}_{N_e}$. This module consists of $N_d$ decoder layers. The structure of each layer is similar to that of the MHA and FFN layers in the transformer encoder. However, there is a difference in that each layer has the source--target attention structure. Therefore, the query is created based on $\bm{h}_{\mathrm{rsa}}$, whereas the key and value are created based on $\bm{h}_{N_e}$. The output is $\bm{h}_{N_d} \in \mathbb{R}^{(k + I + 2) \times d_{\mathrm{dec}}}$, where $d_{\mathrm{dec}}$ denotes the size of each layer.

Finally, the prediction probability of the generated word $p(\hat{\bm{y}}_{t+1,j}) \in \mathbb{R}^{N_v}$ is given by $p(\hat{\bm{y}}_{t+1, j}) = \mathrm{softmax}(f_{\mathrm{gen}}(\bm{h}_{N_d})),$ where $N_v$ denotes a vocabulary size and $f_{\mathrm{gen}}(\cdot)$ denotes the computation by the first fully connected, GELU, LN, and final fully connected layers.
%

The global loss function $\mathcal{L}$ is defined as:
\begin{align}
\mathcal{L} &= \lambda_{\mathrm{ce}} \mathcal{L}_{\mathrm{CE}}(y_{t+1}, p(\hat{\bm{y}}_{t+1})) +  \lambda_{\mathrm{iwp}} \mathcal{L}_{\mathrm{iwp}}(\bm{y}_{t+1, 1},  p(\hat{\bm{y}}_{t+1, 1})) \nonumber \\ &+  \lambda_{\mathrm{corr}} \mathcal{L}_{\mathrm{corr}} + \lambda_{\mathrm{mse}} \mathcal{L}_{\mathrm{MSE}}(\bm{x}_{t+1}, \bm{z}_{t+1}), \nonumber
\vspace{-2mm}
\end{align}
where $\mathcal{L}_{\mathrm{MSE}}(\cdot, \cdot)$  and $\mathcal{L}_{\mathrm{CE}}(\cdot, \cdot)$ denote the mean square error and the cross entropy loss functions, respectively. Here, $\lambda_{\mathrm{ce}}$, $\lambda_{\mathrm{corr}}$, $\lambda_{\mathrm{mse}}$ and $\lambda_{\mathrm{iwp}}$ are hyperparameters. $\mathcal{L}_{\mathrm{corr}}$ penalizes the case where $\hat{y}_{t+1}$ describes an event that occurs before or after $t$. $\mathcal{L}_{\mathrm{iwp}}$ penalizes the case where the predicted word $y^*_{t+1, 1}$ is incorrect. This is defined as $\mathcal{L}_{\mathrm{iwp}}(y_{t+1, 1}, p(\hat{\bm{y}}_{t+1, 1})) = \gamma_{\mathrm{iwp}} \mathcal{L}_{\mathrm{CE}}(y_{t+1, 1},  p(\hat{\bm{y}}_{t+1, 1})),$ where $\gamma_{iwp} = 1/W$. 
The parameter $W$ denotes the number of appearances of the words in the training set.
%
%
%
\vspace{-6mm}
\section{Experiments}
\vspace{-3mm}
\label{exp}
\subsection{Dataset}
\vspace{-2mm}
\begin{table*}[tb]
\centering
\caption{
Quantitative comparison and ablation studies. The best scores are in bold.
}
\begin{adjustbox}{width=\textwidth}
\begin{tabular}{@{}lccccccccc@{}}
\toprule
           & \multicolumn{4}{c}{YouCook2-FC}            & \multicolumn{1}{l}{} & \multicolumn{4}{c}{BILA-caption}           \\ \cmidrule(lr){2-5} \cmidrule(l){7-10} 
Methods      & BLEU4$\uparrow$ & METEOR$\uparrow$ & ROUGE-L$\uparrow$ & CIDEr-D$\uparrow$  &
& BLEU4$\uparrow$ & METEOR$\uparrow$ & ROUGE-L$\uparrow$ & CIDEr-D$\uparrow$  \\ \midrule

MART~\cite{lei2020mart} & $6.85{\scriptscriptstyle \pm0.18}$    & $14.24{\scriptscriptstyle \pm0.07}$  & $\bm{30.80}{\scriptscriptstyle \pm0.21}$    & $20.86{\scriptscriptstyle \pm1.07}$ &                      & $19.01{\scriptscriptstyle \pm0.74}$     & $21.02{\scriptscriptstyle \pm0.45}$  & $30.30{\scriptscriptstyle \pm0.64}$    & $37.33{\scriptscriptstyle \pm4.37}$ \\ \midrule

Ours (w/o RSA)
& $6.70{\scriptscriptstyle \pm0.36}$    
& $14.14{\scriptscriptstyle \pm0.50}$ 
& $30.18{\scriptscriptstyle \pm0.18}$
& $21.26{\scriptscriptstyle \pm2.83}$  &
& $20.37{\scriptscriptstyle \pm0.36}$
& $22.04{\scriptscriptstyle \pm0.19}$
& $40.67{\scriptscriptstyle \pm0.48}$
& $44.65{\scriptscriptstyle \pm4.89}$ \\


Ours (w/o Transformer Decoder)
& $6.68{\scriptscriptstyle \pm0.13}$    
& $14.09{\scriptscriptstyle \pm0.15}$ 
& $30.16{\scriptscriptstyle \pm0.31}$
& $19.75{\scriptscriptstyle \pm1.41}$  &
& $21.08{\scriptscriptstyle \pm1.62}$
& $22.39{\scriptscriptstyle \pm0.84}$
& $40.92{\scriptscriptstyle \pm1.32}$
& $45.05{\scriptscriptstyle \pm6.72}$ \\

\bf{Ours (RFCM)}
& $\bm{7.03}{\scriptscriptstyle \pm0.15}$    
& $\bm{14.53}{\scriptscriptstyle \pm0.09}$ 
& $30.49{\scriptscriptstyle \pm0.21}$
& $\bm{21.32}{\scriptscriptstyle \pm1.09}$  &
& $\bm{21.74}{\scriptscriptstyle \pm1.02}$
& $\bm{22.74}{\scriptscriptstyle \pm0.57}$
& $\bm{41.44}{\scriptscriptstyle \pm0.86}$
& $\bm{49.61}{\scriptscriptstyle \pm8.02}$ \\ \bottomrule

\end{tabular}
\end{adjustbox}
\label{tab:results}
\vspace{-6mm}
\end{table*}
In the experiment, we evaluated our model on the YouCook2-FC and BILA-caption datasets. 
The YouCook2-FC dataset is a dataset for the future captioning task. Generally, in cooking, the next procedure is determined based on the previous procedure. We built the YouCook2-FC dataset based on the YouCook2 dataset~\cite{ZhXuCoAAAI18}.
We set the number of samples included in the training, validation, and test sets to 7435, 1569, and 3035, respectively.


The BILA-caption dataset was newly built to evaluate future captioning models that describe likely collisions in object placement tasks. Fig.~\ref{fig:eye_catch} shows a typical sample from of the BILA-caption dataset. To build the dataset, we extended SIGVerse~\cite{inamura2013development}, which was used in the World Robot Summit 2018 Partner Robot Challenge/Virtual Space competition~\cite{WRS2018}. In the simulator, a DSR placed a randomly selected everyday object (e.g., a bottle or can) in the center of one of five types of furniture (e.g., a table or shelf). Each sample was annotated with a statement explaining the situation that occurred as a result of the robot placement action.
The dataset consists of 1K videos and 1K english captions. The total and average lengths of the videos are 2.2 h and 8 s, respectively. In the dataset, each clip was given a sentence explaining dangerous events (e.g., collision events) and their causes. The vocabulary size is 245 words. We also set the number of samples included in the training, validation, and test sets to 800, 100, and 100, respectively.
%
%
%

\vspace{-5mm}
\subsection{Experimental setup}
\vspace{-2mm}
We preprocessed clips in the datasets as follows. For the YouCook2-FC dataset, we trimmed the videos using the given start and end times. For the BILA-caption dataset, we trimmed the videos so that the clip starts when the arm of the DSR starts to move and ends when one of the following two conditions have been met: a collision event has occurred and more than 4 s have passed since the start time.
Each clip was first converted to 0.6 and 8 fps for the YouCook2-FC and BILA-caption datasets, respectively. For the YouCook2-FC dataset, we used the procedure shown in \cite{ging2020coot}.
For the BILA-caption dataset, we used S3D~\cite{miech19endtoend} pretrained on the Howto100m dataset~\cite{miech19howto100m} to obtain 512-dimensional features. Then, we used a fully connected layer to obtain 384-dimensional features. 

The experimental setup was as follows: the optimizer, learning rate, batch size, and number of epochs were Adam ($\beta_1$: 0.9, $\beta_2$: 0.999), 1.0e-4, 16, and 25, respectively. For each module, $N_e$, $d_{\mathrm{enc}}$, $N_h$, $N_d$, $d_{\mathrm{dec}}$, $N_r$, and $d_{\mathrm{rsa}}$ were 3, 384, 12, 3, 384, 2, 384, respectively. For the loss function, $\lambda_{\mathrm{ce}}$, $\lambda_{\mathrm{iwp}}$, $\lambda_{\mathrm{corr}}$, $\lambda_{\mathrm{mse}}$ were 30, 1.0, 0.1, 0.005, and 10, respectively. For $\mathcal{L}_{\mathrm{iwp}}$, we handled the words that appear more than $n_{\mathrm{th}}$ times. We set $n_{\mathrm{th}}$ to 30. We also sets $W$ for the YouCook2-FC and BILA-caption datasets to 3000 and 1000, respectively. The number of trainable parameters of RFCM and the number of multiply-add operations are 3.1M and 540M, respectively.

The training was conducted on a machine equipped with an NVIDIA Tesla V100 SXM2 with 16 GB of GPU memory, 240 GB RAM, and an Intel Xeon Gold 6148 processor.  It took 6.2 and $1.6 \times 10^{-1}$ h to train our model on the YouCook2-FC and BILA-caption datasets, respectively. Similarly, inference took $1.7 \times 10^{-2}$ and $4.9 \times 10^{-2}$ s/sample on the YouCook2-FC and BILA-caption datasets, respectively.
As a condition for early stopping, we used the generalization described in~\cite{prechelt1998automatic}.
\vspace{-6mm}
\subsection{Quantitative results}
\vspace{-2mm}
We compared RFCM with Memory-Augmented Recurrent Transformer (MART \cite{lei2020mart}). We selected MART as the baseline because it is a representative method for video captioning tasks and can be applied to future captioning tasks. Table~\ref{tab:results} shows the quantitative results on the YouCook2-FC and BILA-caption datasets. The mean and standard deviation were computed on five experimental runs.


The evaluation of the generated sentences was based on several standard metrics for video captioning tasks: BLEU4~\cite{papineni2002bleu}, ROUGE-L~\cite{lin2004rouge}, METEOR~\cite{banerjee2005meteor}, and CIDEr-D~\cite{vedantam2015cider}. The primary metric was CIDEr-D.

First, we compared the models on the YouCook2-FC dataset. The table shows that the CIDEr-D score was improved by 0.46 points. The BLEU4 and METEOR scores were also improved by 0.18 and 0.29 points, respectively.

Next, we compared the models on the BILA-caption dataset. The table shows that the CIDEr-D score was drastically improved by 12.28 points. The other metrics scores were also improved. These results indicate that the RFCM generated sentences more appropriately than the baseline. 
\vspace{-6mm}
\subsection{Qualitative results}
\vspace{-2mm}

\begin{figure}[t]
    \centering
    \includegraphics[height=35mm]{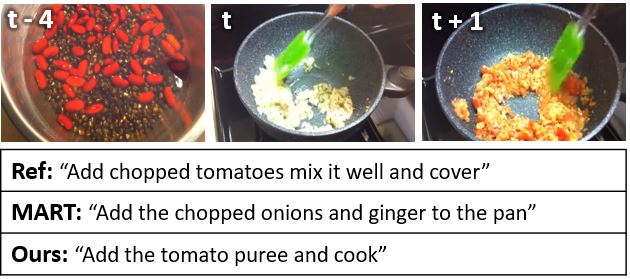}
    \vspace{-5.5mm}
    \caption{\small  The successful examples on the YouCook2-FC dataset. Top figures show events. The bottom table shows a reference sentence and sentences generated by the baseline and our methods.}
    \vspace{-3mm}
    \label{fig:you-suc}
\end{figure}

\begin{figure}[t]
    \centering
    \includegraphics[height=45mm]{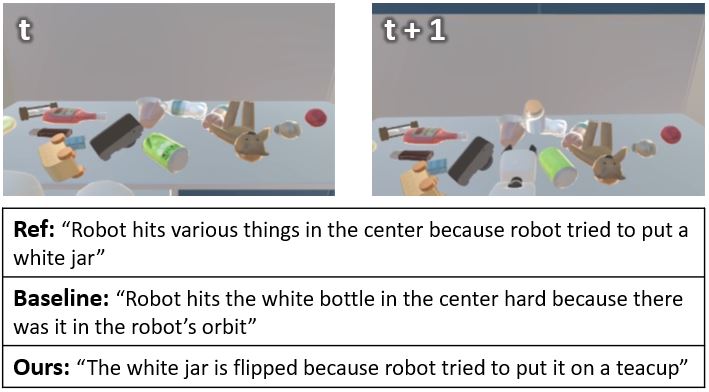}
    \vspace{-5mm}
    \caption{\small  The successful examples on the BILA-caption dataset.}
    \vspace{-6mm}
    \label{fig:bila-suc}
\end{figure}

Figs.~\ref{fig:you-suc}-\ref{fig:bila-suc} show the qualitative results on the YouCook2-FC and BILA-caption datasets. In the figures, the events are shown in chronological order. Some events are omitted due to space limitations. 

The sample in Fig.~\ref{fig:you-suc} illustrates a successful case on the YouCook2-FC dataset. In this example, the object added to the pan was ``chopped tomatoes.'' The baseline method incorrectly described it as ``the chopped onions and ginger.'' In contrast, our method appropriately described it as ``tomato puree.'' This result indicates that our method was able to appropriately predict the next step and generate a caption.

Similarly, the sample in Fig.~\ref{fig:bila-suc} illustrates a successful case on the BILA-caption dataset. In this example, the grasped object was ``the white bottle'' and the collided object was ``the camera.'' The baseline method incorrectly described the collided object as ``a black teapot.'' In contrast, our method appropriately described them as ``a white jar'' and ``the camera,'' respectively. This result indicates that our method could appropriately describe the characteristics of the objects.

\vspace{-2mm}

\vspace{-3mm}
\subsection{Ablation study}
\vspace{-2mm}
We conducted ablation studies for each module. Table~\ref{tab:results} quantitatively shows the mean and standard deviation of five experimental runs. We investigated which module contributed the most to the performance improvement using two ablation conditions: (a) w/o RSA, where we used the standard MHA layer~\cite{vaswani2017attention} instead of the RSA layer, and (b) w/o transformer decoder, where we removed the transformer decoder.

Comparing the results obtained under conditions (a) and (b) with the results of the proposed method on the YouCook2-FC dataset, the CIDEr-D score decreased by 0.06 and 1.57 points, respectively. This indicates that the transformer decoder contributed the most to the performance improvement. 

Similarly, comparing results obtained under the conditions (a) and (b) with the results of the proposed method on the BILA-caption dataset, the CIDEr-D score decreased by 4.96 and 4.56 points, respectively. This indicates that the RSA layer contributed the most to the performance improvement. 

\vspace{-6mm}
\section{Conclusions}
\vspace{-4mm}
In this paper, we focused on the future captioning task, which is a task to generate a description about a future event. Specifically, we proposed a future captioning model for daily tasks.

The main contributions of this paper are as follows:
\vspace{-2mm}
\begin{itemize}
  \setlength{\parskip}{-0.2mm}
  \setlength{\itemsep}{-0.5mm}
 \item We proposed the RFCM, a crossmodal language generation model that can generate a description about a future event. 
 \item The RSA Encoder is introduced to extract the relationships between events more effectively than the conventional self-attention in transformers.
 \item The RFCM outperformed the baseline method on two datasets, BILA-caption and YouCook2-FC.
\end{itemize} 
\vspace{-4mm}

\bibliographystyle{IEEEbib}
{\small \bibliography{reference}}

\end{document}